# Conflict and Surprise: Heuristics for Model Revision

**Kathryn Blackmond Laskey**
Department of Systems Engineering and C³I Center
George Mason University
Fairfax, VA  22030

## Abstract

Any probabilistic model of a problem is based on assumptions which, if violated, invalidate the model. Users of probability based decision aids need to be alerted when cases arise that are not covered by the aid's model. Diagnosis of model failure is also necessary to control dynamic model construction and revision. This paper presents a set of decision theoretically motivated heuristics for diagnosing situations in which a model is likely to provide an inadequate representation of the process being modeled.

## 1  INTRODUCTION

Building a model for an inference problem involves constructing and reasoning within a restricted universe of propositions relevant to the inference problem at hand. Following Savage (1954), I term this restricted universe a *small world* (actually, a small *set* of possible worlds). The small world must obviously include those propositions of direct inferential interest. It also includes certain other propositions which bear on the propositions of interest and about which information may be available, either directly or indirectly. A model for the inference problem specifies relationships (logical or probabilistic) between propositions in the small world, and inference rules for revising beliefs as information is obtained.

The small world includes those propositions represented explicitly in the inference system. But a model of the relationships between these propositions often depends on a background context that is not explicitly represented. The model may make inferences that are seriously in error if these background assumptions are violated. For example, a medical diagnosis system may confidently misdiagnose a patient who is actually suffering from a disease it does not know about. A threat assessment system will be deceived by electronic interference that puts "ghost targets" on the radar screen if the behavior of the interference device is not represented in its knowledge base. A navigation system may go awry if unforseen weather conditions impact the performance of its sensors.

No modeler can hope to cover all possibilities that might arise. If the model is a good one, the situations it cannot handle should be rare. But it is important to be able to recognize such situations when they arise. Even when a system cannot revise its own model of the situation, it can alert the user that its model may be inadequate in the current situation. As research in dynamic model construction matures, model failure indicators will provide an important component of a control strategy for model construction and revision. It may be necessary or desirable not to explore some search paths during network construction, or to prune parts of the network when hypotheses become improbable or nearly independent of the hypotheses of interest. But an improbable hypothesis may become more probable as more evidence is observed. In such situations, a trigger is needed to alert the system that it may be necessary to explore search paths that were initially ignored, or to bring back pruned parts of the network.

## 2  BACKGROUND

A small world for an inference problem can be represented as a vector $\underline{X}$ of *propositional variables*. Each variable $X_i$ can take on values in the set $X_i \in \{x_{i1}, \ldots, x_{ik_i}\}$. Let $\underline{X}_e$ denote a subvector of $\underline{X}$ whose values have been observed; these are called the *evidence* variables. Let $\underline{X}_t$ denote a subvector of *target* variables, or variables whose values have not been observed, but are of direct interest. Denote the remaining *unobserved* variables as $\underline{X}_u$. Unobserved variables may become evidence variables if their values are observed at a future time, but at present their values are of interest only because of their relationship to the target variables. Assume the variables are ordered so that $\underline{X} = \{\underline{X}_t, \underline{X}_e, \underline{X}_u\}$. The goal of inference is to draw conclusions about the values of the target variables $\underline{X}_t$ given observed values $\underline{x}_e$ for the evidence variables.

A probabilistic model for the small world assigns a probability distribution over the variables in the small world. The *assessed* probability for an assignment of values to the variables is denoted by $P^a(\underline{x})$. Inference within the model consists of conditioning the assessed probability distribution on the observed values for the evidence variables:

$$P^a(\underline{x}_t, \underline{x}_u \mid \underline{x}_e) = \frac{P^a(\underline{x})}{P^a(\underline{x}_e)}, \text{ or}$$

$$P^a(\underline{x}_t \mid \underline{x}_e) = \sum_{\underline{x}_u} P^a(\underline{x}_t, \underline{x}_u \mid \underline{x}_e). \qquad (1)$$

A model is at best an approximation of what is being modeled. The distribution $P^a(\cdot)$ is assessed relative to some assumed context; if the assumptions are violated, then the model no longer applies. If $P^a(\cdot)$ is a good approximation, one should feel confident that the assumptions underlying the model are at least approximately correct. Intuitively, the approximation is good if $P^a(\underline{x}_t \mid \underline{x}_e)$ is nearly correct for most $\underline{x}_e$.

This intuitive notion of a good approximation is far from precise, and glosses over some important questions. What does it mean for $P^a(\underline{x}_t \mid \underline{x}_e)$ to be "nearly correct?" What should be the definition of "most $\underline{x}_e$?" I return to these questions later, when I formalize the idea of approximating a model. For now, an intuitive understanding of the quality of an approximation should suffice.

Until recently, research on automated probabilistic inference has focused on computationally efficient methods for computing (1). Much less consideration has been given to the issue of deciding whether (1) is an adequate representation of the data generating process. The statistical community has devoted more attention to this issue, and there is a large literature on the theory of statistical hypothesis testing. However, the statistical hypothesis testing paradigm assumes an explicit, well formulated alternative model against which the current model is tested. In artificial intelligence applications, the purpose of model diagnosis is to initiate search for an alternative model. Requiring explicit representation and computation of the alternative model prior to model diagnosis defeats the purpose of the entire enterprise.

Research on model diagnosis from an AI perspective is beginning to receive more attention. There is general agreement that suspicion should be aroused when a combination of evidence items occurs that was initially assessed to be highly improbable: that is, when $P^a(\underline{x}_e)$ = LOW.[1] The problem with this idea is defining LOW.

When there are many uncertain evidence items, the probability of any one combination of values is bound to be quite small. Thus, the definition of a low-probability evidence combination must be relative to the other evidence combinations that might have occurred.

Habbema (1976) suggests identifying a set of "surprising" observations. An observation in the surprising set triggers extra diagnostic attention for the case in question. According to Habbema's definition, each observation identified as surprising must be less probable than all observations not considered surprising; and the total probability of the set of surprising observations must be less than some threshold $\alpha$. Laskey (1990) suggests comparing $P^a(\underline{x}_e)$ with the expected value $E[P^a(\underline{X}_e)]$ of the probability of the evidence. These suggested approaches do provide measures of relative improbability, but they appear to be computationally intractable for the inference network models common in the literature on uncertainty in AI.

Jensen et al. (1990) suggest a tractable indicator of *conflict* between items of evidence. Their conflict measure can be written:

$$c_J = \log_2\left(\frac{P^a(x_{e1}) \cdots P^a(x_{ek})}{P^a(x_{e1}, \ldots, x_{ek})}\right). \qquad (2)$$

where the $x_{ei}$ are the observed values of the components of the evidence vector $\underline{X}_e = (X_{e1}, \ldots, X_{ek})$. This measure is easy to compute using any of the currently popular evidence propagation algorithms. Both the numerator and denominator of (2) can be produced as a natural byproduct of evidence accumulation, if each node stores its original prior probability distribution $P^a(X_i)$ along with its current conditional probability distribution $P^a(X_i \mid \underline{x}_e)$. The numerator of (2) is just the product of the prior probabilities of the observed values of the evidence variables. The denominator is also straightforward to compute as follows. Assume that evidence items $x_{e1}, \ldots, x_{ek-1}$ have been observed, that $P(x_{e1}, \ldots, x_{ek-1})$ has been computed, and that the evidence absorption algorithm has resulted in the computation of a revised distribution for $X_{ek}$, namely $P^a(X_{ek} \mid x_{e1}, \ldots, x_{ek-1})$. Now, when $x_{ek}$ is observed, the required joint probability can easily be computed: $P^a(x_{e1}, \ldots, x_{ek}) = P^a(x_{ek} \mid x_{e1}, \ldots, x_{ek-1}) P^a(x_{e1}, \ldots, x_{ek-1})$.

Jensen et al. justify $c_J$ heuristically: they simply assert that one would expect the observed evidence to have higher probability than the product of the joint probabilities. Intuitively, if $x_{e1}$ has been observed, it might be reasonable to expect to observe values for $X_{e2}$

---

[1] This recommendation appears to violate the likelihood principle, a central tenet of Bayesian theory. According to the likelihood principle, the likelihood of data that might have been observed is irrelevant; all that matters is what was observed. But the likelihood principle applies to the agent's "true" model. The likelihood of unobserved data may indeed be relevant to the issue of whether the current approximation remains tenable.

**Figure 1: $C_J$ Positive With Probability .55**

P(x,y)

|    | y1    | y2    | y3    |       |
|----|-------|-------|-------|-------|
| x1 | .1125 | .11   | .11   | .3325 |
| x2 | .1125 | .11   | .11   | .3325 |
| x3 | .11   | .1125 | .1125 | .335  |
|    | .335  | .3325 | .3325 |       |

P(x)P(y)

|    | y1    | y2    | y3    |       |
|----|-------|-------|-------|-------|
| x1 | .1114 | .1106 | .1106 | .3325 |
| x2 | .1114 | .1106 | .1106 | .3325 |
| x3 | .1122 | .1114 | .1114 | .335  |
|    | .335  | .3325 | .3325 |       |

$c_J$ given (x,y)

|    | y1     | y2     | y3     |
|----|--------|--------|--------|
| x1 | -.0413 | .0073  | .0073  |
| x2 | -.0413 | .0073  | .0073  |
| x3 | .0289  | -.0413 | -.0413 |

that are made more likely by the observation of $x_{e1}$, that is, values $x_{e2}$ for which

$$P(x_{e2} \mid x_{e1}) > P(x_{e2}) \text{, or}$$

$$P(x_{e1}, x_{e2}) = P(x_{e2} \mid x_{e1})P(x_{e1})$$

$$> P(x_{e1})P(x_{e2}) \, . \qquad (3)$$

But the example in Table 1 demonstrates that it is not necessarily ture that (3) is satisfied most of the time. In this example, there is a .55 chance that $c_J$ is greater than zero, i.e., there is a better than even chance that the product of the marginal probabilities exceeds the joint probability.

Although it may be probable that $c_J$ is greater than zero, I will show in the next section that very large values of $c_J$ are highly improbable. Furthermore, $c_J$ is never positive in expected value. Its expectation is given by:

$$E[c_J] = -\sum_{\underline{x}_e} P^a(\underline{x}_e) \log_2 \left( \frac{P^a(\underline{x}_e)}{P^a(x_{e1}) \cdots P^a(x_{ek})} \right) . \qquad (4)$$

This quantity is the negative of the information theoretic distance from the probability distribution $P^a(\cdot)$ to the distribution $P^i(\cdot)$, in which all the $X_{ei}$ are independent, but their marginal distributions are the same as their marginal distributions under $P^a(\cdot)$. That is, the *expected value* of $c_J$ is a measure of how closely the probability distribution $P^i(\cdot)$ approximates the assessed distribution $P^a(\cdot)$. A well-known result from information theory states that (4) is never positive and is equal to zero only if the distributions $P^a(\cdot)$ and $P^i(\cdot)$ are identical (Kullback, 1959). In other words, the expected value of $c_J$ is more negative the greater the nonindependency among evidence items--that is, the less well $P^i(\cdot)$ approximates $P^a(\cdot)$.

When $c_J$ is positive, the probability $P^i(\underline{x}_e)$ of $\underline{x}_e$ under the independence model is greater than the probability $P^a(\underline{x}_e)$ under the assessed distribution. In other words, a model in which the evidence items are independent fits the observed evidence better than does the assessed distribution. Now, the whole modeling exercise was based on the assumption that the $X_{ei}$ were related to $\underline{X}_t$, and therefore to each other. That is, combinations of evidence items characteristic of a particular $\underline{x}_t$ should tend to occur together. When the independence model fits better than the assessed distribution, it is an indication that the particular combination $\underline{x}_e$ of observed values is characteristic of no $\underline{x}_t$ under the current model.

The next section generalizes $c_J$ to a family of "model suspicion" measures based on comparing how well $P^a(\cdot)$ fits the evidence relative to an alternative model. Unlike formal statistical hypothesis testing, the alternative model is generally not taken seriously as a candidate model for the data generating process. Rather, its purpose is to alert a system or user that the current model may be inadequate given the current situation.

## 3 THEORETICAL FRAMEWORK

It is now time to develop a formal framework for model approximation and model failure diagnosis. Assume that the small world $\underline{X}$ is embedded within a larger world $\underline{W} = (\underline{X}', \underline{Y})$. The vector $\underline{X}'$ represents the same variables as $\underline{X}$, but each may have additional outcomes in the larger world that are not represented in the small world. That is, $X'_i \in \{x_{i1}, \ldots, x_{ik_i}, x_{i(k_i+1)}, \ldots, x_{ir_i}\}$, where only the first $k_i$ values are also possible values for $\underline{X}$. The vector $\underline{Y}$ corresponds to variables that are not explicitly represented in the small world. Assume that there is some probability distribution $P(\cdot)$ on the larger world which $P^a(\cdot)$ is intended to approximate.

Interpreting the distribution $P(\cdot)$ raises philosophical issues which lie beyond the scope of this paper, but a few words on how I interpret $P(\cdot)$ are appropriate. My main concern is with an automated reasoning system which computes belief values for the target variables $\underline{X}_e$ based on any evidence $\underline{x}_e$ that has been observed to date. Such systems are usually engineered by assessing probabilities from some expert or experts in the domain about which the system reasons. In this context, $P^a(\cdot)$ represents a probability distribution assessed from the expert, who has restricted attention to the small world $\underline{X}$. I assume that $P^a(\cdot)$ approximates some distribution $P(\cdot)$ over $\underline{W}$, in the sense that $P^a(\cdot)$ is obtained from $P(\cdot)$ by conditioning on the small world $\underline{X}$. This does not necessarily mean that $P(\cdot)$ exactly represents the expert's beliefs over $\underline{W}$, or even that the expert has well-defined and coherent "true" probabilities over $\underline{W}$. What I do assume is that if the expert were to make the effort to assess beliefs over the expanded world $\underline{W}$, that this would result in the distribution $P(\cdot)$, and that this distribution is a more accurate representation of the expert's beliefs than the assessed distribution $P^a(\cdot)$. In other words, if computation and assessment burden were not an issue, and if a second system were built using the distribution $P(\cdot)$, the expert would feel more satisfied with the output of the second system than with the output of the system based on the model $P^a(\cdot)$. It is also assumed that the expert thinks it is unlikely that the assumptions underlying the assessments are violated--that is, the expert thinks the small world is probable.

A somewhat different interpretation is appropriate for systems designed for dynamic model revision. Here it is assumed that the full distribution $P(\cdot)$ was assessed from the expert, but is represented only implicitly in the system's knowledge base. The system has explicitly constructed just a small portion of this large implicit joint probability distribution. The assumptions underlying the model the system constructs amount to conditioning on the small world.

Let the proposition q represent the assumptions underlying the assessed distribution. The proposition q includes the restriction of the values of $\underline{X}'$ to $\underline{X}$, as well as some assumptions about the variables $\underline{Y}$:[2]

$$q = (\bigwedge_i X_i' \in \{x_{i1},...,x_{ik}\}) \wedge (\bigvee_{\underline{y}_j} \underline{Y} = \underline{y}_j). \qquad (5)$$

$P^a(\cdot)$ was assessed under the assumption that q was the case. That is, for any $\underline{x}$ in the small world $\underline{X}$:

$$P^a(\underline{x}) = P(\underline{x} \mid q) = \sum_{\underline{y}} P(\underline{x},\underline{y} \mid q). \qquad (6)$$

If q is not the case, the model $P^a(\cdot)$ is not appropriate, and should be replaced by:

$$P^o(\underline{x}) = P(\underline{x} \mid \neg q) = \sum_{\underline{y}} P(\underline{x},\underline{y} \mid \neg q). \qquad (7)$$

Neither the alternative model nor the probability $P(q)$ is assessed explicitly. However, I assume that the context q is assumed because it is thought to be probable, i.e., $P(q) = 1-\varepsilon$, where $\varepsilon$ is small. The model $P(\cdot)$, restricted to the variables $\underline{X}'$, can be written:

$$P(\underline{x}) = (1-\varepsilon)P^a(\underline{x}) + \varepsilon P^o(\underline{x}) \qquad (8)$$

(where $P^a(\underline{x})$ is understood to be equal to zero if one of the $x_i$ is outside the range of $X_i$). Because $\varepsilon$ is very small, $P^a(\underline{x})$ provides a good approximation to the correct joint probability $P(\underline{x})$.

To summarize, I assume that the small world model $P^a(\cdot)$ over $\underline{X}$ is obtained by a combination of *conditioning* on assumptions thought to be probable (i.e., conditioning on q) and *marginalizing* on variables not of direct interest (i.e., summing over values of $\underline{Y}$).

The goal of inference is to estimate $P(\underline{x}_t \mid \underline{x}_e)$. Because the conditioning operator is nonlinear, there is no guarantee that the approximation error will remain small as evidence is absorbed. The relationship between the actual and estimated posterior probabilities given $\underline{X}_e = \underline{x}_e$ can be seen from the following expression:

$$P(\underline{x}_t \mid \underline{x}_e) =$$

$$\frac{(1-\varepsilon)P^a(\underline{x}_t,\underline{x}_e) + \varepsilon P^o(\underline{x}_t,\underline{x}_e)}{(1-\varepsilon)P^a(\underline{x}_e) + \varepsilon P^o(\underline{x}_e)}$$

$$= (1-\varepsilon^*)P^a(\underline{x}_t \mid \underline{x}_e) + \varepsilon^* P^o(\underline{x}_t \mid \underline{x}_e), \qquad (9)$$

where

$$\varepsilon^* = \frac{\varepsilon P^o(\underline{x}_e)}{(1-\varepsilon)P^a(\underline{x}_e) + \varepsilon P^o(\underline{x}_e)}.$$

Comparing (8) and (9), it is clear that the "true" prior and posterior probabilities have the same form: both are weighted averages of the assessed model probability $P^a(\cdot)$ and the unknown alternative model probability $P^o(\cdot)$. The prior probability $\varepsilon$ of the assessed model is replaced in (9) by $\varepsilon^*$, the posterior probability of the assessed model given that the evidence variables $\underline{X}_e$ take on values $\underline{x}_e$.

Note that

---

[2] This formulation might seem to preclude assumptions about the relationship between variables in $\underline{X}$ (e.g., that $X_i$ and $X_j$ are independent). However, one could [psot that $X_i'$ and $X_j'$ are independent conditional on $Y_k$, where $Y_k$ has high probability.

$$\frac{\varepsilon^*}{1-\varepsilon^*} = \frac{\varepsilon}{(1-\varepsilon)} \cdot \frac{P^o(\underline{x}_e)}{P^a(\underline{x}_e)} . \quad (10)$$

Thus, the approximation error becomes large when $P^o(\underline{x}_e)$ is so much larger than $P^a(\underline{x}_e)$ that it swamps the difference in magnitude between $\varepsilon$ and $1-\varepsilon$. The first term in (10) is called the *prior odds ratio*; the second term is called the *likelihood ratio* of the data given the two models being compared.

It would seem straightforward, then, to decide when a model is no longer a good approximation to the observed data. Simply compare the probability of the observations under the approximate model to the probability under the alternative model, and initiate model revision when the ratio of these probabilities becomes too small. But a moment's thought reveals that this will not do: the reason for adopting the approximation in the first place was to avoid the expense of explicitly constructing a detailed model of the many improbable contexts in which $P^a(\cdot)$ does not apply. In other words, you need to have already performed model revision in order to have $P^o(\underline{x}_e)$ available for computing (10).

However, it may be possible to formulate "straw models," which capture some of the expert's intuitions about how the model could go wrong but are computationally much simpler than the fully specified alternate model. The independence model described in Section 2 is just such a model. The following theorem shows that a straw model is unlikely to fit much better than the assessed model in cases for which the assessed model applies.

**Theorem 1:** Let $P^a(\cdot)$ and $P^s(\cdot)$ be probability distributions over $\underline{X}$. Define the index of surprise at evidence $\underline{x}_e$ under $P^a(\cdot)$ relative to $P^s(\cdot)$ as:

$$c_S = \log_2\left(\frac{P^s(\underline{x}_e)}{P^a(\underline{x}_e)}\right) . \quad (11)$$

Let $\pi_K$ be the probability under $P^a(\cdot)$ that $c_S$ is greater than K. Then $\pi_K < 2^{-K}$.

**Proof:**

$$1 = \sum_{\underline{x}_e} P^s(\underline{x}_e) \geq \sum_{\frac{P^s(\underline{x}_e)}{P^a(\underline{x}_e)} > 2^K} \frac{P^s(\underline{x}_e)}{P^a(\underline{x}_e)} P^a(\underline{x}_e)$$

$$\geq \sum_{\frac{P^s(\underline{x}_e)}{P^a(\underline{x}_e)} > 2^K} 2^K \cdot P^a(\underline{x}_e) = 2^K \pi_K .$$

Therefore, $\pi_K < 2^{-K}$ .  ∎

A trivial consequence of Theorem 1 is that high values of conflict are *a priori* unlikely when the assessed model is considered probable. That is, *any* alternative model specified *a priori* is unlikely to fit the data much better than the assessed model.[3]

**Corollary 1:** If $\underline{X}_e$ is distributed according to $P(\underline{x}_e) = (1-\varepsilon)P^a(\underline{x}_e) + \varepsilon P^o(\underline{x}_e)$, where $P^o(\cdot)$ is a distribution not necessarily the same as $P^s(\cdot)$, then the probability that $c_S$ is less than K exceeds $(1 - \varepsilon)(1 - 2^{-K})$.

The trick is to specify an alternate model that is likely to fit the data better than the assessed model when $P^a(\cdot)$ does *not* apply. That is, we would like for $c_S$ to be large when $\varepsilon^*$ is large (or at least in the most probable situations in which $\varepsilon^*$ is large). To construct a model revision indicator, then, the modeler thinks carefully about situations in which the model might not apply, and about how the predictions of the model might fail when this happens. The modeler then formulates a straw model $P^s(\cdot)$ which is computationally simple but captures some important features of situations in which the assessed model is likely to fail.

Consider for example the conflict indicator $c_J$. The evidence variables $\underline{x}_e$ are predictors of the target variables. They should therefore be expected to be marginally dependent--certain patterns of values of the $x_{e_i}$ are likely to occur together because they indicate a particular target vector $\underline{x}_t$; other combinations of values are unlikely because they are unlikely given *any* target vector. Without formulating an alternate model in detail, the modeler can still use situations in which the independence model fits the data better than the assessed model to diagnose possible problems with the assessed model.

## 4  REBUTTALS

A general theoretical framework for heuristic model revision indicators has been presented. In summary, the system is assumed to reason under a current model, which is regarded as an approximation to some more accurate model which may or may not be represented implicitly in the system's knowledge base. The approximate model is conditioned on an assumed context which the system regards as highly probable. If the assessed model is an accurate model of the phenomenon in question, it would be expected to fit the data better than other candidate models. A heuristic model revision indicator can be constructed by building an alternate "straw model" which is not necessarily taken seriously as a model of the phenomenon in question, but which is likely to fit better than the assessed model in some class of situations which violate the assumed context.

---

[3] Of course, one can always specify after the fact a model that predicted the observed data exactly. When there are many uncertain evidence items, the observed evidence will have a very low probability, and this "20-20 hindsight" model will fit much better than the assessed model. However, the *a priori* probability that exactly this model would fit the data was extremely low.

I have already considered one example of a "straw model" which gives rise to Jensen et al.'s conflict indicator $c_J$. In this section, I relate another proposed model revision strategy (Laskey, 1990) to the framework proposed in Section 3.

A *Bayesian network* for the probability model $P^a(\cdot)$ is a directed acyclic graph in which each node corresponds to a variable $X_i$ and the directed arcs encode direct conditional dependencies. Suppose that $N$ is a Bayesian network for $P^a(\cdot)$ and $X_{i_1}, \ldots, X_{i_k}$ is an ordering of the variables such that all predecessors of $X_{i_j}$ in $N$ are also predecessors in the ordering. Then $X_{i_j}$ is independent of all preceding $X_{i_m}$ given its immediate predecessors in $N$. The direct parents of the node $X_i$ are denoted by the vector $\underline{X}_{p(i)}$.

A probability model and associated Bayesian network on a set $\underline{X}$ of variables can be completely characterized by its *node models*. A node model for variable $X_i$ is the tuple

$$\mathcal{M}_i = (X_i, \underline{X}_{p(i)}, P^a(X_i \mid \underline{X}_{p(i)}))  \quad (12)$$

consisting of the variable, its direct parents in $N$, and a set of conditional probability distributions for the node, one for each combination of values for its parent variables.

In previous work (Laskey, 1990; see also Laskey, Cohen and Martin, 1989), I have suggested associating *rebuttals* with each node model.[4] A rebuttal is a proposition expressing a condition under which the assessed probability distribution for the node model does not apply. Rebuttals make explicit the contextual assumptions underlying the model: in the terminology of Section 3, the truth of a rebuttal implies $\neg q$. Thus, the system can monitor the rebuttal propositions and trigger model revision when a rebuttal is observed to be true.

However, the system may not have the resources to check rebuttals routinely, or obtaining information about some rebuttals may be costly in time or other resources. Sometimes it might be desirable to associate a general "*ceteris non paribus*" rebuttal with a node model. That is, one allows for the possibility that the node model does not apply, but does not explicitly model the circumstances under which the assumptions are violated (unless the general rebuttal becomes sufficiently probable to warrant such effort).

Let $R_i$ be a variable taking on values $R_i \in \{t_i, f_i\}$, where $t_i$ means that at least one of the rebuttals to $\mathcal{M}_i$ is true, and $f_i = \neg t_i$ means that all rebuttals are false. The assessed node model distribution assumes $f_i$. That is, $P^a(X_i \mid \underline{X}_{p(i)}) = P(X_i \mid \underline{X}_{p(i)}, f_i)$. To completely specify a model including the rebuttal variable $R_i$ would mean specifying two additional distributions: the distribution of $X_i$ given its parents and $t_i$, and the distribution of $t_i$. Additional complications may arise: $t_i$ might imply that $X_i$ depends directly on other nodes in addition to $\underline{X}_{p(i)}$, and $R_i$ might depend on other nodes or other rebuttals (some condition may invalidate more than one node model). The network might become computationally intractable if these complexities were included explicitly (in fact, assuming them away may have been a computationally motivated approximation).

It may be useful to specify a straw model which does not cover all these complexities, but might be expected to fit better than the assessed model when one of the rebuttals is true. Consider the following simplified model. First, a rebuttal is assumed to "break the link" between a node and its parents, so that:

$$P^s(X_i \mid \underline{X}_{p(i)}, t_i) = P^s(X_i \mid t_i) . \quad (13)$$

That is, given $t_i$, $X_i$ is conditionally independent of all the nodes above it in $N$, and has the distribution $P^s(X_i \mid t_i)$. Second, the rebuttals for different node models are assumed to be marginally independent of all other nodes in the network (i.e., $R_i$ has no predecessors in $N$ or among the other $R_j$).

If the rebuttals are included explicitly in the inference network, the posterior probability $P^s(R_i \mid \underline{x}_e)$ is computed as part of evidence propagation. If the general rebuttal $R_i$ becomes probable, it may flag the system to build a more detailed model of the conditions under which the current node model for $X_i$ is not valid.

However, explicit representation of rebuttals doubles the number of nodes in the network.[5] It is desirable to find a computationally simple method to determine when a rebuttal $R_i$ has become probable. To determine the probability of $R_i$ given evidence, one computes the posterior odds ratio by multiplying the likelihood ratio by the prior odds ratio:

$$\frac{P^s(t_i \mid \underline{x}_e)}{P^s(f_i \mid \underline{x}_e)} = \frac{P^s(\underline{x}_e \mid t_i)}{P^s(\underline{x}_e \mid f_i)} \times \frac{P^s(t_i)}{P^s(f_i)} . \quad (14)$$

When the likelihood ratio is large, the evidence $\underline{x}_e$ increases the posterior probability of $t_i$ relative to its prior probability. This likelihood ratio can be rewritten as:

---

[4] The term rebuttal is due to Toulmin (e.g., Toulmin et al., 1984); its use in this context is due to Marvin Cohen (cf., Cohen, Laskey and Ulvila, 1987).

[5] No loops are added to the network, so algorithms to find loop cutsets or clique trees need work no harder.

$$\frac{P^s(\underline{x}_e \mid t_i)}{P^s(\underline{x}_e \mid f_i)} =$$

$$\frac{\displaystyle\sum_{(x_i, \underline{x}_{p(i)})} P^s(\underline{x}_e \mid x_i, \underline{x}_{p(i)}) P^s(x_i \mid t_i) P^s(\underline{x}_{p(i)})}{\displaystyle\sum_{(x_i, \underline{x}_{p(i)})} P^s(\underline{x}_e \mid x_i, \underline{x}_{p(i)}) P^a(x_i \mid \underline{x}_{p(i)}) P^s(\underline{x}_{p(i)})} \quad . \quad (15)$$

Note that the first and last terms in the summands are the same in numerator and denominator, and that these terms are independent of $R_i$. If $R_i$ is the only rebuttal in the model, (15) can be reexpressed as:

$$\frac{\displaystyle\sum_{(x_i, \underline{x}_{p(i)})} P^a(\underline{x}_e \mid x_i, \underline{x}_{p(i)}) P^s(x_i \mid t_i) P^a(\underline{x}_{p(i)})}{\displaystyle\sum_{(x_i, \underline{x}_{p(i)})} P^a(\underline{x}_e \mid x_i, \underline{x}_{p(i)}) P^a(x_i \mid \underline{x}_{p(i)}) P^a(\underline{x}_{p(i)})}$$

$$= \frac{\displaystyle\sum_{(x_i, \underline{x}_{p(i)})} P^a(\underline{x}_e \mid x_i, \underline{x}_{p(i)}) P^a(x_i, x_{p(i)}) \frac{P^s(x_i \mid t_i)}{P^a(x_i \mid x_{p(i)})}}{P^a(\underline{x}_e)}$$

$$= \sum_{(x_i, \underline{x}_{p(i)})} P^a(x_i, \underline{x}_{p(i)} \mid \underline{x}_e) \frac{P^s(x_i \mid t_i)}{P^a(x_i \mid x_{p(i)})} \quad . \quad (16)$$

When there are rebuttals for other nodes as well, (16) is only approximate. It is a close approximation if the posterior probabilities of all rebuttals other than $R_i$ are small.

The sum (16) will be large when the evidence $\underline{x}_e$ makes probable values of $(x_i, \underline{x}_{p(i)})$ for which the straw model probability $P^s(x_i \mid t_i)$ is much larger than the assessed probability $P^a(x_i \mid \underline{x}_{p(i)})$. This might happen if $x_i$ is a very atypical value of $X_i$ given the values $\underline{x}_{p(i)}$ for the parent variables, but the evidence makes both $x_i$ and $\underline{x}_{p(i)}$ very probable.

Clique tree algorithms automatically keep track of the joint probabilities of the $(x_i, \underline{x}_{p(i)})$; other algorithms can be modified to do so. However, computation of (16) generally amounts to about as much work as explicitly modeling $R_i$. Computation can be reduced by pre-selecting a subset of $(x_i, \underline{x}_{p(i)})$ for which $P^s(x_i \mid t_i)$ is much larger than $P^a(x_i \mid \underline{x}_{p(i)})$, and monitoring the posterior probability for only the selected subset of values for $X_i$ and its parents.

## 5  RARE CASES

I have suggested that model revision should be considered when low probability evidence is observed. But even if the model is correct, things which it assigns low probability are occasionally expected to happen. Is there any way to distinguish between rare cases for which the model is correct and cases not covered by the model?

Sometimes the conflict can be explained by some rare hypothesis which is covered by the model. That is, there may be some variable $X_k$ for which the value $x_{k_H}$ would make the observed evidence $\underline{x}_e$ highly probable. The probability of the evidence can be written as:

$$P^a(\underline{x}_e) = P^a(\underline{x}_e \mid x_{k_H}) P^a(x_{k_H})$$
$$+ P^a(\underline{x}_e \mid \bar{x}_{k_H}) P^a(\bar{x}_{k_H}) \; .$$

This can be very small even when $P^a(\underline{x}_e \mid x_{k_H})$ is high if $x_{k_H}$ was initially thought to be a very improbable value for $X_k$. If such a rare hypothesis could explain the conflict, the value of the conflict indicator $c_S$ will typically be decreased by independent evidence for $x_{k_H}$. That is, one finds an observable variable $X_f$ for which the assessed probability of $x_{f_H}$ is high given $X_k = x_{k_H}$ but low given $X_k \neq x_{k_H}$. Observing $X_f$ may resolve the issue. Precisely speaking, the value $(\underline{x}_e, x_{f_H})$ was assessed to be much more probable relative to other values of $(\underline{X}_e, X_f)$ than was the value $\underline{x}_e$ relative to other values of $\underline{X}_e$. Unless this is also true of the straw model, the conflict will be reduced upon observing $\underline{x}_{f_H}$.

In any case, it is generally wise to single out for special attention cases which occur but were initially assessed to be highly improbable. It may be that, being rare, they were not thought deserving of as close modeling attention. Or it may have been that a heuristic model construction algorithm pruned parts of the network that were independent of the variables of interest given the falsity of the improbable hypotheses. These modeling decisions may need to be reexamined if the hypotheses become more probable than originally thought.

## 6  DISCUSSION

If probabilistic reasoning methods are to be used on problems where flexible network reconfiguration is necessary, indicators must be constructed that tell the system when its current model appears to be inadequate. Even for static models, it is important to alert users to situations which the system's model was not designed to handle. An exact decision theoretic calculation would require actually revising the model to decide whether model revision is necessary. This is clearly infeasible. This paper presented a class of theoretically justified heuristic model revision indicators. The idea is to construct a computationally simple alternate model which, although not a tenable model for the phenomenon, is likely to fit the evidence better than the current model if the current model is incorrect. Model revision is initiated when the likelihood ratio of the evidence given the alternate model becomes unacceptably large.

**Acknowledgements**

This work was supported by a grant from the Virginia Center for Innovative Technology to the Center of Excellence in Command, Control, Communications and Intelligence at George Mason University.